# Variational Program Inference

**Georges Harik**

Harik Shazeer Labs
444 Ramona
Mountain View, CA 94043
georges@gmail.com

**Noam Shazeer**

Google Research
1600 Amphitheatre Pkwy.
Palo Alto, CA 94301
noam@google.com

**Abstract**

We introduce a framework for representing a variety of interesting problems as inference over the execution of probabilistic model programs. We represent a "solution" to such a problem as a guide program which runs alongside the model program and influences the model program's random choices, leading the model program to sample from a different distribution than from its priors. Ideally the guide program influences the model program to sample from the posteriors given the evidence. We show how the KL-divergence between the true posterior distribution and the distribution induced by the guided model program can be efficiently estimated (up to an additive constant) by sampling multiple executions of the guided model program. In addition, we show how to use the guide program as a proposal distribution in importance sampling to statistically prove lower bounds on the probability of the evidence and on the probability of a hypothesis and the evidence. We can use the quotient of these two bounds as an estimate of the conditional probability of the hypothesis given the evidence. We thus turn the inference problem into a heuristic search for better guide programs.

## Problem Specification

Given partial observations of a complicated system governed by known or unknown probabilistic rules, we would like to automatically reason about the likely state of hidden parts of the system.

### Systems with Known Rules

We model our system as a program in a general purpose programming language. We call this the **model program**. We will be agnostic to what programming language we are using. We will only insist that it be deterministic except for a `choose` function which takes a probability distribution as an argument and returns a random choice from it. The model program thus defines a probability distribution $P(x)$ over **execution paths** x. If over the course of an execution path x of the model program, the choose function is called n times with distributions $(P_1, P_2 ... P_n)$ and the randomly chosen values are $(c_1, c_2 .. c_n)$ respectively, then the probability of that execution path is

$$P(x) = \prod_i P_i(c_i).$$

We are interested in the conditional expected value $E(h(x)|e)$, where **h** is some function of the execution path, and **e** is some evidence such that we can easily compute $P(e|x)$ for any x.

Our programming language needs to include constructs for specifying $P(e|x)$ and $h(x)$. The model program reports $P(e|x)$ as the product of all calls to a function: `evidence`. This is a particularly convenient in that the evidence function can be passed boolean values which are interpreted as 0 or 1. For example, we could represent an observation that the grass is wet with the call `evidence(grass_wet)`. If `grass_wet` is false, $P(e|x)$ is multiplied by 0, and if `grass_wet` is false, $P(e|x)$ is left unchanged.

The value of the hypothesis $h(x)$ is defined as the final value of a global variable *h*. We may not have a hypothesis, and may only be interested in sampling runs of the program given the evidence. In this case, *h* might not be set or used.

**Example 1**: Three fair dice are rolled and it is observed that their sum is 7. What is the probability that the first die rolled was a 5? The following model program (shown in pseudocode) might encode that problem.

```
die1 := choose(uniform(1..6))
die2 := choose(uniform(1..6))
die3 := choose(uniform(1..6))
*h* := (die1 == 5)
evidence (die1 + die2 + die3 == 7)
```

### Modeling Systems with Unknown Rules

A system with unknown rules is in fact a special case of a system with known rules if we model the unknown rules as being randomly generated by known rules.

In particular, we include in our model program a **random program generator**. A random program generator is a function which generates a random program and executes it. The exact code of the random program generator will depend on the programming language and our arbitrary choice of probability distribution. It turns out that our arbitrary choice of random program generator is not critical, since if there were a better random program generator, our random program generator would have some finite chance of generating it, and so we underestimate the probability of generating any program by at most that constant factor. It could be a large constant factor, but the larger our problems and the more evidence we have, the less significant that constant factor will be.

**Example 2**: an unknown function f returns 9 when given 3 and 16 when given 4. What does f(5) return?

```
<random program generator>
evidence (f(3) == 9)
evidence (f(4) == 16)
print (f(5))
```

In this example, in order to give the evidence high probability, the random program generator will need to make choices that define f as some function which has a good chance of returning 9 when called with 3 and 16 when called with 4.

When using a random program generator, there are, of course limitations on what the generated program can do. For instance, in the above example, the generated program is not allowed to redefine the evidence function or the equality operator, or to exit the model program early, or to read the values 9 and 16 out of the input program. The details of how to protect against this will depend on our choice of programming language.

**Example 3**: designing an agent to act in a known environment
Our model represents a known environment which makes calls to an agent for a decision. The agent's decision function will be left up to a random program generator. There is evidence that the agent accomplishes a goal. For example, The environment could be "blocks world", chess versus a known program as an opponent, a traveling salesman problem, etc.

We end up reasoning about what likely algorithms will make the agent likely to accomplish its goal. In such a model, we may need to be careful not to allow the agent's decision function to read or modify parts of the environment that it is not allowed to. Again, how we do this is language dependent.

**Example 4**: explaining the hidden structure in any large data set.

```
<random program generator>
evidence(d == <our data set>)
```

The data set could be the text of wikipedia, the set of undeciphered Linear A inscriptions, Tycho Brahe's astronomical observations, the text of the web, etc. This is the problem that most excites us. In our opinion, it's a fundamental problem of science - trying to find the most likely generative explanation of the world. We hope that our inference system, when given a set of physical observations, will hypothesize the laws of physics in order to explain the observations in the most likely manner. Likewise, when asked to explain the text of the web, we hope that our system will hypothesize the existence of a physical universe and how it may have given rise to the text.

## Inference

By "inference", we mean that we want to compute the expected value $E(h(x)|e)$ of some hypothesis h which is a function of the execution path x of the model, given the evidence e. (If we want the probability of a binary predicate, we can represent that as the expected value of the 0/1 indicator function for that predicate).

(1)   $E(h(x)|e) = \Sigma_X P(x|e)\, h(x)$
       $= \Sigma_X P(x)\, P(e|x)\, h(x) / P(e)$
       $= \Sigma_X P(x)\, P(e|x)\, h(x) / \Sigma_X P(x)\, P(e|x)$

In general, this is utterly intractable to compute, requiring one to enumerate over all execution paths of the model.

As mentioned above, for a particular execution path x, $P(x)$ and $P(e|x)$ are easy to compute. $P(e)$ and $P(x|e)$ are hard to compute, since they appear to also require a complete enumeration.

To introduce a little notation, let's say that
   X is the set of execution paths
   $P_X$ is the prior probability distribution over execution paths. $P(x)$ is synonymous with $P_X(x)$.
   $P_{X|e}$ is the posterior probability distribution over execution paths given the evidence. So $P_{X|e}(x) = P(x|e)$.

### Inference by sampling

It would be tremendously useful if we could quickly sample from the posteriors $P_{X|e}$. If we could do so, we could estimate $E(h(x)|e)$ by averaging $h(x)$ over many execution paths drawn from the posteriors.

### Exact Sampling by Rejection

We can sample slowly from the posteriors by rejection - sampling from the priors $P_X$ by running the model program and discarding a sample x with probability $(1-P(e|x))$. This is in general too slow, since the expected number of tries to produce a sample goes up with the inverse likelihood of the evidence, and the evidence is likely to be highly improbable in most interesting problems. We will therefore concentrate on sampling approximately from the posteriors.

### Single Point "Sampling" (Maximum Likelihood

Search)

One way to sample approximately from the posteriors is to heuristically search for the most likely single execution path we can find given the evidence, and return it every time. Some advantages of this technique are:
1. It's simple
2. It can yield incremental results
3. The results are measurable.
4. It can be easily parallelized across computers and across search heuristics.
5. The results can be good for probability distributions that are concentrated around a single execution path.

The downside of such a maximum likelihood search is that it performs poorly when the posterior distribution is spread out. For instance, in our example of the three dice which are observed to add to 7, there are many possible execution paths, all with identical posterior probabilities. This method would just pick one and be sure of it. Furthermore, this method discriminates against execution paths which cause more random numbers to be thrown, even if the results of those numbers are not used.

## Guided Sampling (Variational Program Inference)

Our strategy for sampling will be to run the model program, but to bias the probability distribution on each one of its choices so that instead of drawing from the priors $P_X$, the biased model program instead draws from a distribution which approximates the posteriors $P_{X|e}$.

We do this by means of a guide program G which runs alongside the model program and influences its random choices. Every time the main program calls the choose command with a prior probability distribution $P_C$, the guide program computes and substitutes in its own distribution $G_C$, and the choice c is chosen from $G_C$ instead. Running the model program with the guide program alongside it samples from a guide-influenced distribution $G_X$ over execution paths. Our goal is to find a guide program G such that $G_X$ approximates $P_{X|e}$.

For example, this is example 1 from above with a good guide program represented inline in angle brackets <>:

```
die1 := choose(uniform(1..6),
    <guide: {1:1/3, 2:4/15, 3:1/5, 4:2/15, 5:1/15}>)
die2 := choose(uniform(1..6),
             <guide: uniform(1..6-die1)>)
die3 := choose(uniform(1..6),
             <guide: {7-(die1+die2):1.0}>)
*h* := (die1 == 5)
evidence (die1 + die2 + die3 == 7)
```

We say that the guide program runs alongside the model program, but in practice, it will be more convenient to have it consist of code segments which are attached inline to the model program. The one caveat is that the guide program may not affect the execution of the model program in any way except to provide alternate probability distributions for choose calls.

We will find a good guide program by heuristically searching over the space of all guide programs for one such that $G_X$ approximates $P_{X|e}$. We will worry about search heuristics later, but the more pressing problem for now is how to tractably judge for a particular G how well $G_X$ approximates $P_{X|e}$.

## Evaluating a Guide Program by Free Energy

The measure of similarity we use is the Kulllbeck-Leibler divergence between $G_X$ and $P_{X|e}$, which is defined as
$$D_{KL}(G_X \| P_{X|e}) = \Sigma_x G(x) \log(G(x) / P(x|e))$$
The KL divergence measures how quickly a wary observer can accrue evidence that you are sampling from $G_X$ instead of from $P_{X|e}$. Smaller values of KL divergence indicate more similar distributions.

By an amazing trick, the KL divergence becomes tractable to estimate if we will add the constant term $-\log(P(e))$. Thus, we define the free energy of G as:
$$F(G, P, e) := D_{KL}(G_X \| P_{X|e}) - \log(P(e))$$
$$= \Sigma_x G(x) \log(G(x) / P(x|e)) - \log(P(e))$$
$$= \Sigma_x G(x) \log(G(x) / (P(x|e) * P(e)))$$
$$= \Sigma_x G(x) \log(G(x) / P(x,e))$$
$$= \Sigma_x G(x) \log(G(x) / (P(x) * P(e|x)))$$
$$= \Sigma_x G(x) ( \log(G(x) / P(x)) - \log(P(e|x)) )$$

Lo and behold, we can estimate this last quantity by sampling from $G_X$ multiple times and computing the average value of $\log(G(x) / P(x)) - \log(P(e|x))$.
We call the quantity we are averaging the one-run free energy an execution path
$$F(G, P, e, x) := \log(G(x) / P(x)) - \log(P(e|x))$$

This technique of searching for approximate posteriors which minimize the free energy is related to variational methods of probabilistic inference, where the guide program is the variational parameter. So we call the technique of searching for an optimal guide "Variational Program Inference". This can be read as either "Variational Program-Inference" or "Variational-Program Inference", since we are running inference over programs, and the variational parameter itself is a program.

Our technique differs from the typical use of variational probabilistic inference in that we estimate the free energy via sampling, while typically, the free energy is computed analytically. We assume that our problems will be too complex to allow for such an analytic computation.

Conveniently, we can separate the one-run free energy for a run into one term for each call to the choose function or the evidence function. $F(G, P, E, x)$ is the sum of

- For each call to the choose function, $\log(G_C(c)/P_C(c))$, where c is the value that was chosen, $P_C(c)$ is the probability of that choice according to the distribution provided by the model, and $G_C(c)$ is the probability of that choice according to the distribution provided by the guide.
- For each call `evidence(p)` to the evidence function, $-\log(p)$

So we have a very easy way of computing the one-run free energy. Furthermore, we can assign credit/blame to the particular parts of the program that contribute to the one-run free energy. This could prove useful in directing our optimization efforts.

If $G_X = P_{X|e}$, we call G a **perfect guide**. For example, the guide shown above for example 1 happens to be a perfect guide. If G is a perfect guide, here are some interesting facts:

- The free energy $F(G, P, e) = -\log(P(e))$, the smallest possible value.
- The one run free energy is the same for every possible execution path x. $F(G, P, e, x) = -\log(P(e))$
  (the converse is not true - a constant one-run free energy does not imply a perfect guide)
- Across different runs, the constant one-run free energy separates into terms in very different ways.
- For every call to the choose function, the guide distribution for that choice $G_C$ is equal to the posterior distribution for that choice $P_{C|e}$

### Guided Sampling with Rejection

So far, we have only dealt with guides that always work. In practice, for a given guide program, the model program or the guide program could occasionally crash or exhaust our computing resources, or the one-run free energy could sometimes be very low. This isn't a reason to dismiss that guide program entirely. Say we reject the runs where either the program crashed or where the one-run free energy exceeded some threshold specified in the guide program. Accepting only the successful runs induces its own distribution $G'_X$ over execution paths. Let the acceptance rate $A(G)$ be the fraction of runs that are accepted. For any accepted execution path x,
$$G'_X(x) = G_X(x) / A(G)$$
$$F(G', P, e, x) = F(G, P, e, x) - \log(A(G))$$

So to estimate the free energy of G', we compute the average value of $F(G, P, e, x)$ over the non-rejected samples, and subtract the log of the observed acceptance rate.

### Other Utility Metrics for Guide Programs

In addition to favoring guide programs with lower free energy, we may want to favor guide programs which make life easier for us in other ways - most notably by costing us fewer computing resources to sample from. For example, a guide program that just echoes the model program's priors and then rejects at the end with probability $(1-P(e|x))$ has an optimal free energy, but is very expensive to sample from. So we define a utility on guide programs
  $U(G) = F(G, p, e) + k *$ (average time to successfully extract a sample from G) + other terms
and we optimize for that instead. k here is a constant representing our level of impatience.

### Pros and Cons of Guided Sampling

Guided sampling can be seen as an extension of maximum likelihood search (searching for the most likely execution path). The two methods are identical if we restrict our search for guides to "deterministic" guides that always provide single-point distributions for every choice.

As such, guided sampling shares some of the the advantages of maximum likelihood search. In particular
  1. It can yield incremental results
  2. The results are measurable.
  3. It can be easily parallelized across computers and across search heuristics.
On the negative side:
  1. Guided sampling is more complicated than maximum likelihood search.
  2. It requires repeated sampling to produce and verify results.

## Importance Sampling

There may be something better we can do with a guide program. Say we are trying to compute the conditional probability of a hypothesis $h(x)$ given the evidence e (rather than just wanting to sample from $P_{X|e}$ *per se*). Instead of a 0/1 boolean, we can let $h(x)$ be any positive function and we will try to determine its conditional expected value given e. Recall equation 1:

(1) $\quad E(h(x)|e) = \Sigma_X P(x) P(e|x) h(x) / \Sigma_X P(x) P(e|x)$

This is the quotient of two sums of the form
$$F = \Sigma_X P(x) f(x)$$
where $f(x)$ is an easily computable non-negative function. These sums can be very time-consuming to compute exactly. To approximate them, we turn to importance sampling. Importance sampling will allow us to statistically prove lower bounds on the two sums. While this proves nothing about their quotient, we believe this can often yield a good estimate.

Say we have a guide program which induces a probability distribution G(x).

$$F = \Sigma_x P(x)f(x) \geq \Sigma_{x|G(x)>0} G(x)P(x)f(x)/G(x) = \langle f(x)P(x)/G(x) \rangle_G$$

That is, our sum is greater than or equal to the expected value under G of $f(x)P(x)/G(x)$, with equality coming when $G(x)$ is positive for all x such that $f(x)P(x)$ is non-zero. We can therefore sample from G and use the samples $f(x)P(x)/G(x)$ to find statistical lower bounds for F. Our problem is now that of heuristically finding a G that lets us efficiently prove a good statistical lower bound on F. A test for this is given in [Breth et al.].

We search for a guide algorithm which lets us quickly prove as great a lower bound as we can on F. A trivial guide which sampled from the priors would have the correct mean $\langle f(x)P(x)/G(x) \rangle_G = F$, but the variance could be so great that it could take exponentially long to statistically prove a good lower bound on F. A **perfect guide** here would be one which $f(x)P(x)/G(x) = F$ for all x.

Note that in general, estimation of the numerator and the denominator from the right hand side of equation 1 have different perfect guides.

**Extra Guide Choices**

It might sometimes be easier for the guide to specify extra choices not made by the model program. Say our model is that a monkey types a string of a million random characters, and the evidence is that the text of this paper occurs somewhere in the string. One good guide would first choose a position in the string for this paper to occur. It would then guide the choices of those characters to contain the characters of this paper and leave the rest random. The system so far described forbids us from counting the probability of the choice of starting position in computing G(x). This is because we have no proof that different values of extra choices made by the guide lead to different execution paths of the model program. Thus we could end up underestimating G(x) and hence overestimating F.

To solve this problem, we introduce the idea of model extensions. Before making an extra choice y, the guide program extends the model from a distribution over x to one over (x,y) by specifying a function for the conditional distribution $P_G(y|x)$ in terms of the complete execution path x. The guide program then provides a distribution G(y) and the choice of y is chosen from it. At the end of the execution of the model program, we can compute $P_G(x,y) = P(x)P_G(y|x)$ and $G(x,y) = G(x)G(y)$. We do importance sampling on the distribution $P_G(x,y)$ using $G(x,y)$ as the proposal distribution. $P_G(x,y)$ marginalizes to $P(x)$, so the expectation of any function of x is unchanged.

In the case of the prolific monkey, y is the position at which our paper will appear in the monkey's output. The guide makes G(y) the uniform distribution over all possible starting positions. The guide sets $P_G(y|x)$ to a function of the execution path x that returns the position of the first occurrence of our paper in the output.

We extend this method to making multiple extra choices. The guide program has a command to insert an extra choice $y_i$ at any point in the execution. If the model choices made so far are $(x_1...x_j)$ and the extra guide choices made so far are $(y_1...y_{i-1})$ the guide program specifies a distribution $G(y_i|x_1..x_j,y_1..y_{i-1})$ from which the choice is actually made, and a function from the complete execution path x to a conditional distribution $P_G(y_i|x,y_1...y_{i-1})$. If we consider y to be the sequence of extra choices $(y_1...y_n)$ made over the course of he execution, then we are drawing from G(x,y) and we can at the end compute $P_G(x,y) = P(x)P_G(y_1|x)...P_G(y_n|x,y_1...y_{n-1})$. So, as above, we can do importance sampling over $P_G(x,y)$ using $G(x,y)$ as a proposal distribution. Since $P_G(x,y)$ marginalizes to P(x), we can use this to sample from P(x).

## Acknowledgments

Thanks to Jeremy Bem for introducing us to program induction, Sergey Pankov to importance sampling, and Noah Goodman, Vikash Mansinghka, Daniel Roy for a discussion on extra choices.

## References

Breth, M., Maritz, J. and Williams, E. (1978). *On distribution-free lower confidence limits for the mean of a nonnegative random variable*. Biometrika 65 529-534.

Jordan, M.I., Ghahramani, Z., Jaakkola, T.S. and Saul, L.K. (1999). *An Introduction to Variational Methods for Graphical Models*. In Machine Learning, 37, #2, 183-233.

Pearl, J. (1988) <u>Probabilistic Reasoning in Intelligent Systems: Networks of Plausible Inference</u>, San Mateo, CA: Morgan Kaufmann.